\documentclass{article}

\PassOptionsToPackage{numbers}{natbib}
\usepackage[preprint]{neurips_2021}

\usepackage[utf8]{inputenc} % allow utf-8 input
\usepackage[T1]{fontenc}    % use 8-bit T1 fonts
\usepackage{hyperref}       % hyperlinks
\usepackage{url}            % simple URL typesetting
\usepackage{booktabs}       % professional-quality tables
\usepackage{amsfonts}       % blackboard math symbols
\usepackage{nicefrac}       % compact symbols for 1/2, etc.
\usepackage{microtype}      % microtypography
\usepackage{xcolor}         % colors
\RequirePackage{tabularx}
\usepackage{graphicx}
\usepackage{epsfig}
\usepackage{dcolumn}% Align table columns on decimal point
\usepackage{subcaption,soul}
\newcommand{\etal}{\textit{et al}. }
\usepackage{amssymb}

\title{MetaGraspNet\_v0: A Large-Scale Benchmark Dataset for Vision-driven Robotic Grasping via Physics-based Metaverse Synthesis}

\author{%
   Yuhao Chen\thanks{Equal contribution} \\
   University of Waterloo \\
   \texttt{yuhao.chen1@uwaterloo.ca} \\
   \And
   E. Zhixuan Zeng\footnotemark[1] \\
   University of Waterloo \\
   \texttt{ezzeng@uwaterloo.ca} \\
   \And
   Maximilian Gilles\footnotemark[1] \\
   Karlsruhe Institute of Technology \\
   \texttt{maximilian.gilles@kit.edu} \\
   \And
   Alexander Wong \\
   University of Waterloo \\
   \texttt{alexander.wong@uwaterloo.ca} \\
}

\begin{document}

\maketitle

\begin{abstract}
There has been increasing interest in smart factories powered by robotics systems to tackle repetitive, laborious tasks. 
One particular impactful yet challenging task in robotics-powered smart factory applications is robotic grasping: using robotic arms to grasp objects autonomously in different settings.
Robotic grasping requires a variety of computer vision tasks such as object detection, segmentation, grasp prediction, pick planning, etc. 
While significant progress has been made in leveraging of machine learning for robotic grasping,  particularly with deep learning, a big challenge remains in the need for large-scale, high-quality RGBD datasets that cover a wide diversity of scenarios and permutations. 

To tackle this big, diverse data problem, we are inspired by the recent rise in the concept of metaverse, which has greatly closed the gap between virtual worlds and the physical world.
In particular, metaverses allow us to create digital twins of real-world manufacturing scenarios and to virtually create different scenarios from which large volumes of data can be generated for training models.
In this paper, we present MetaGraspNet: a large-scale benchmark dataset for vision-driven robotic grasping via physics-based metaverse synthesis.
The proposed dataset contains 100,000 images and 25 different object types, and is split into 5 difficulties to evaluate object detection and segmentation model performance in different grasping scenarios.
We also propose a new layout-weighted performance metric alongside the dataset for evaluating object detection and segmentation performance in a manner that is more appropriate for robotic grasp applications compared to existing general-purpose performance metrics.
The MetaGraspNet benchmark dataset will be available open-source on Kaggle\footnote{\url{https://www.kaggle.com/metagrasp/metagraspnetdifficulty1-easy}, \\ \url{https://www.kaggle.com/metagrasp/metagraspnetdifficulty2-medium}, \\ \url{https://www.kaggle.com/metagrasp/metagraspnetdifficulty3-hard1}, \\ \url{https://www.kaggle.com/metagrasp/metagraspnetdifficulty4-hard2}, \\ \url{https://www.kaggle.com/metagrasp/metagraspnetdifficulty5-very-hard}}, with the first phase consisting of detailed object detection, segmentation, layout annotations, and a script for layout-weighted performance metric (\url{https://github.com/y2863/MetaGraspNet}).
\end{abstract}

\section{Introduction}
\label{sec:intro}
% The use of robotic systems to perform repetitive, laborious tasks like handing and sorting objects or managing material flow has gained increased traction over the years.
There has been increasing interest in smart factories powered by robotics systems to tackle repetitive, laborious tasks such as handing and sorting objects or managing the material flow.
One particular impactful yet challenging task in robotics-powered smart factory applications is robotic grasping which involves using robotic arms to grasp objects.
A common robotic grasping scenario found in production system or warehouses includes moving a specific object from one bin to another (order picking).
The seemingly simple task for human is quite complex for the robots to perform, requiring a variety of computer vision tasks such as object detection, segmentation, grasp prediction, pick planning, etc. 
While significant progress has been made in the leveraging of machine learning strategies for robotic grasping \cite{graspnet1billion, contact-graspnet, clear-grasp, ggcnn},  particularly with deep learning, a very big challenge in tackling this problem is the need for large-scale, high-quality RGBD datasets that cover a wide diversity of scenarios and permutations (e.g., different combination of objects, different ordering and orientation of objects, different ways of stacking objects.).  
%current model sizes (large scale)
%(high quality)
%diverse scenarios/permutations, 6DOF, objects per-environment, and difficulties
Many existing grasping datasets \cite{graspnet1billion, roi-gd, xiang2017posecnn, asif2018graspnet, mahler2017dex, yan2018learning} provide large-scale and high image quality data, but  they have simple and similar environment settings, such as objects are placed in a common way without stacking.
Another important attribute current large grasping datasets lack is environment layout on how the objects are positioned and stacked, especially in a cluttered environment.
In scenarios where robotic arms are required to pick a specific item from a cluttered scene, picking an obstructed object before removing obstacles could lead to significant damage as the objects covering it are forced out of the way. 
With environment layout labels, pick planning can be trained more intelligently to avoid object damages \cite{vmrd, graphbased_vmrsn, panda-relation-dataset}.
However, there are only a few datasets \cite{vmrd, panda-relation-dataset, large-scale-relation-grasp} providing layout labels.
Most of the datasets \cite{vmrd, panda-relation-dataset} have limited training value as they lack in data size, depth information, as well as segmentation labels.
In addition, objects should be picked sequentially.
Occluded objects will be revealed once top objects are picked.
Thus, not all the objects in a scene are equally important, and top objects are more important to evaluate.
An object detection and segmentation metric weighted according the environment layout would better reflect the performance of a model for a robotic grasping task.

Motivated to tackle this big, diverse data problem, we are inspired by the recent rise in the concept of metaverse, which has greatly closed the gap between virtual worlds and the physical world.  
In particular, metaverses allow us to create digital twins of real-world manufacturing scenarios and enter these metaverses to virtually create different scenarios from which large volumes of high quality data can be generated for training models.
In this paper, we present MetaGraspNet: a large-scale benchmark dataset for vision-driven robotic grasping via physics-based metaverse synthesis.
This dataset contains 100,000 RGBD images, 11,000 scenes, and 25 classes of objects.
The dataset is split into 5 difficulties to evaluate object detection and segmentation model performance in different grasping scenarios.
In addition, we propose a new layout-weighted performance metric alongside the MetaGraspNet benchmark dataset for evaluating object detection and segmentation performance in a manner that is more appropriate for robotic grasp applications compared to existing general-purpose performance metrics.
The proposed MetaGraspNet benchmark dataset will be available in an open-source form on Kaggle \cite{kaggle}, with the first phase consisting of detailed object detection, segmentation, layout annotations, and a script for layout-weighted performance metric (\url{https://github.com/y2863/MetaGraspNet}).

\begin{figure}

	\centering
	\subfloat[]{\label{fig:rgb}{\epsfig{figure=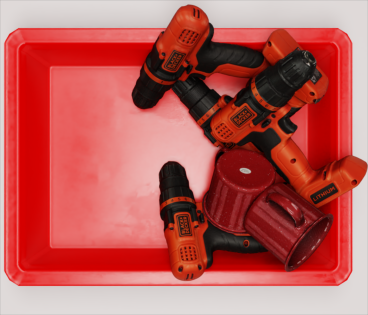,width = 0.22\textwidth}}}
    \,
	\subfloat[]{\label{fig:depth}{\epsfig{figure=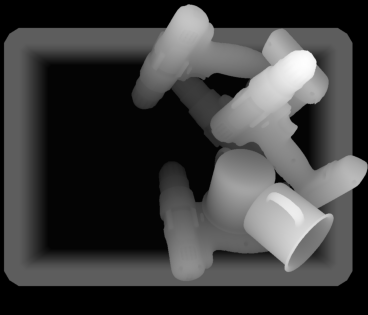,width = 0.22\textwidth}}}
    \,
	\subfloat[]{\label{fig:label}{\epsfig{figure=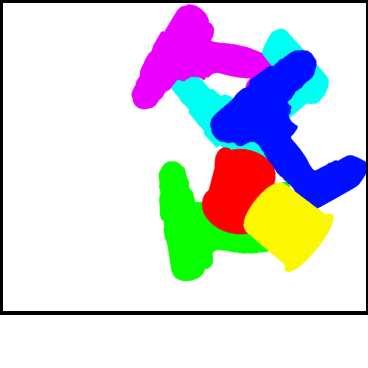,width = 0.21\textwidth}}}
\,
	\subfloat[]{\label{fig:occlusion}{\epsfig{figure=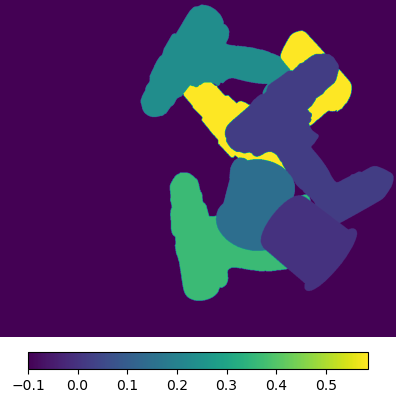,width = 0.21\textwidth}}}

	\caption{Example data in MetaGraspNet benchmark dataset: a) RGB image. b) Depth image. c) Instance annotations. d) Occlusion percentage annotations. Objects are marked with their occlusion percentage, while background is marked with a value of -0.1 .}
	\label{fig:example}
\end{figure}

\section{Dataset}

\subsection{Physics-based Data Synthesis in the Metaverse}
Manually capturing object detection datasets in real-world robotic bin picking environments is intractable in most practical scenarios for a number of key reasons.  First of all, this manual capturing process involves repeatedly setting up physical grasping environment, physically placing different objects into the physical environment, recording images with sensors, and removing objects.  As such, this process is very time consuming, laborious, and unscalable in most practical scenarios as it requires the entire capture process to be manually repeated for each environment and scenario.  
Second, the manual process of placing different objects into different layouts by a human operator also means that the way the objects are arranged in three dimensional space often does not reflect how objects are physically dropped together into a pile during the material handling processes in real-world warehouses or industry related scenarios.
Third, manually labeling the sensor data is very time consuming, therefore static, and cannot keep up with the emerging demand for data needed for training deep neural networks. In \cite{graspnet1billion}, Fang \etal have come up with an intuitive way to overcome the enormous labeling effort for each viewpoint by mounting the camera to the robotic end effector and recording the relative movement between image frames, however their approach still needs precise initial manual annotations for each scene and is restricted to known objects and physical environments with a robotic manipulator.
Image synthesis approaches that generated images based on randomizing object counts, poses, and positions can be used to cut down the data collection time significantly, but creates unrealistic or physically impossible layouts where an object can overlap with other objects in the same spatial location.  As such, the effectiveness of training deep learning models using images generated in this fashion can be very limited for real-world deployment scenarios.  Therefore, a way to generated large-scale benchmark datasets with highly diverse environments and layout permutations for vision-driven robotic grasping in a scalable yet realistic manner is highly desired.

Motivated by this, we take inspiration from the recent rise of metaverses, which are highly immersive virtual environments that facilitates for significant interaction. The significant advancements in metaverses have significantly closed the gap between virtual worlds and the physical world, particularly in physics-based metaverse creation platforms such as the Nvidia Omniverse \cite{omniverse}.   

\begin{figure}[htb]
  \centering
  \includegraphics[width=0.5\linewidth]{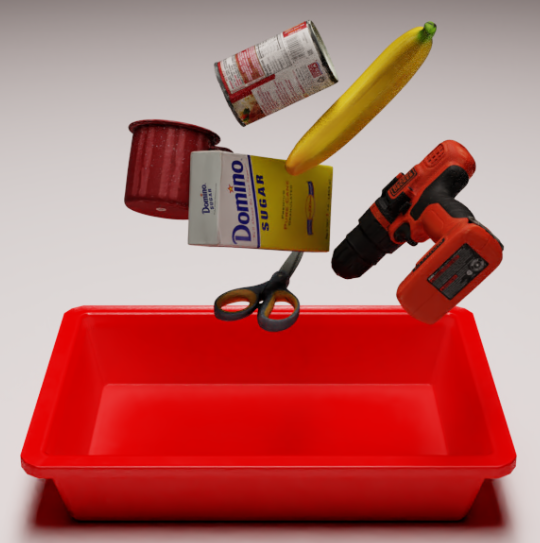}
   \caption{Physics-based data synthesis: Items are dropped into the photorealistic, physics-driven metaverse digital twin of different manufacturing scenarios.}
   \label{fig:drop}
\end{figure}

To create the proposed MetaGraspNet benchmark dataset, we leverage Nvidia Omniverse to create photorealistic, physics-driven digital twins of different real-world manufacturing scenarios. Within these digital twins, we then randomly drop objects under different environment configurations and let the objects interact through physics simulation to ensure the object layouts as captured within the MetaGraspNet dataset are realistic and physically accurate (as shown in Figure \ref{fig:drop}).   Performing the data capturing process in such realistic manufacturing digital twin metaverses enables us to greatly scale in data quantity and diversity beyond what is possible with real-world manual capturing approaches in a very efficient and effective manner, but also allow us to obtain high quality, realistic data that mimics real-world physical scenarios well beyond what is possible with image synthesis approaches.

\subsection{Object Layout Label}
\label{sec:layout-label}

\begin{figure}[]
  \centering
  \includegraphics[width=0.4\linewidth]{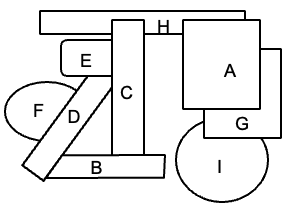}
  \includegraphics[width=0.5\linewidth]{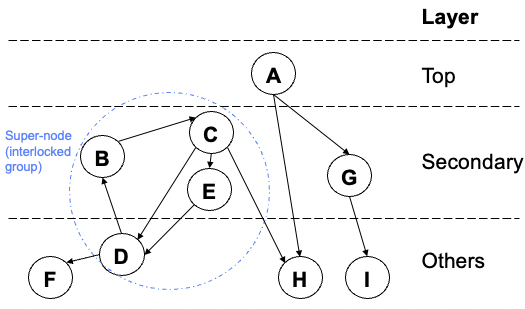}
   \caption{Example of how a graph of objects is categorized into the different layers (Top, secondary, others). Each graph edge represents an instance of an object covering the other.}
   \label{fig:layers-graph-and-figure}
\end{figure}

In addition to the typical semantic mask labels, we propose three more labels to characterize the object layouts.

\begin{figure*}[htb]

	\centering
	\subfloat[]{\label{fig:lvl1}{\epsfig{figure=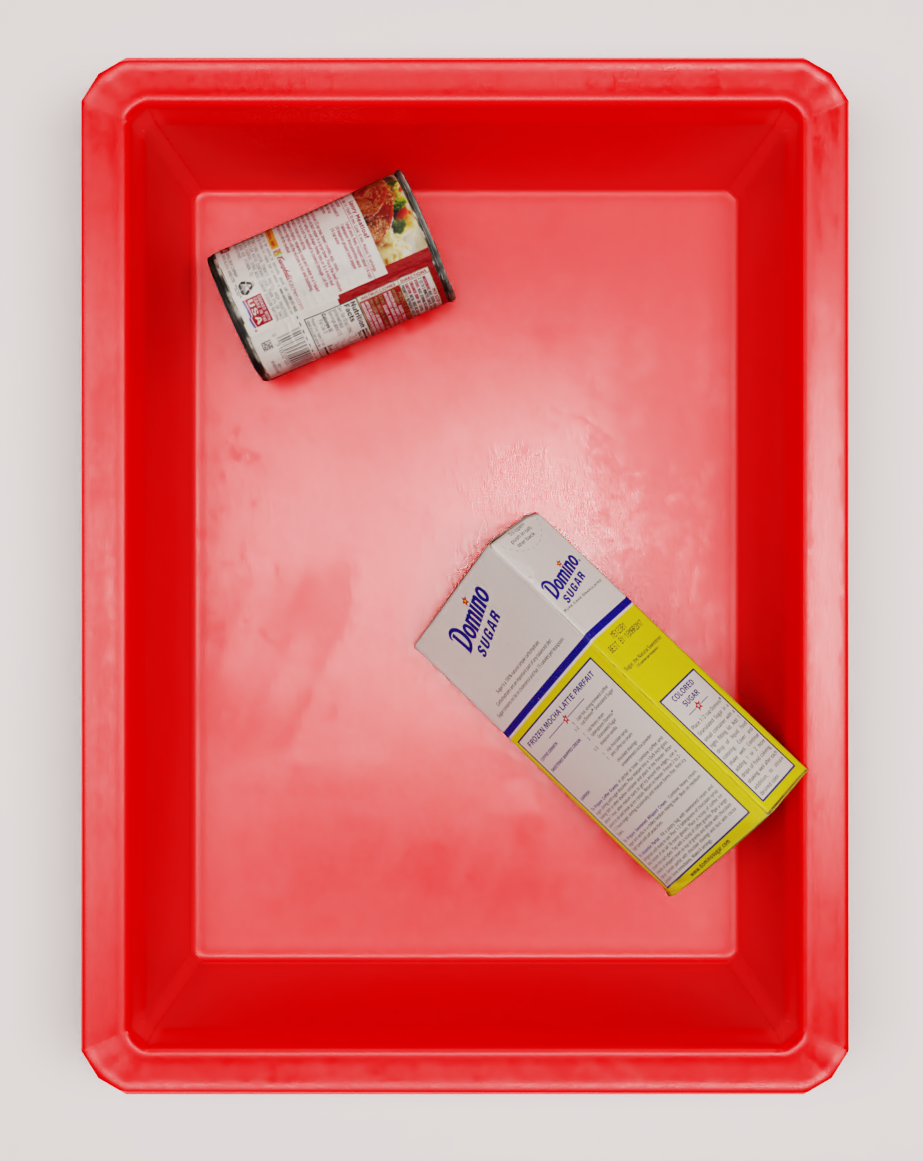,width = 0.19\textwidth}}}
    \,
	\subfloat[]{\label{fig:lvl2}{\epsfig{figure=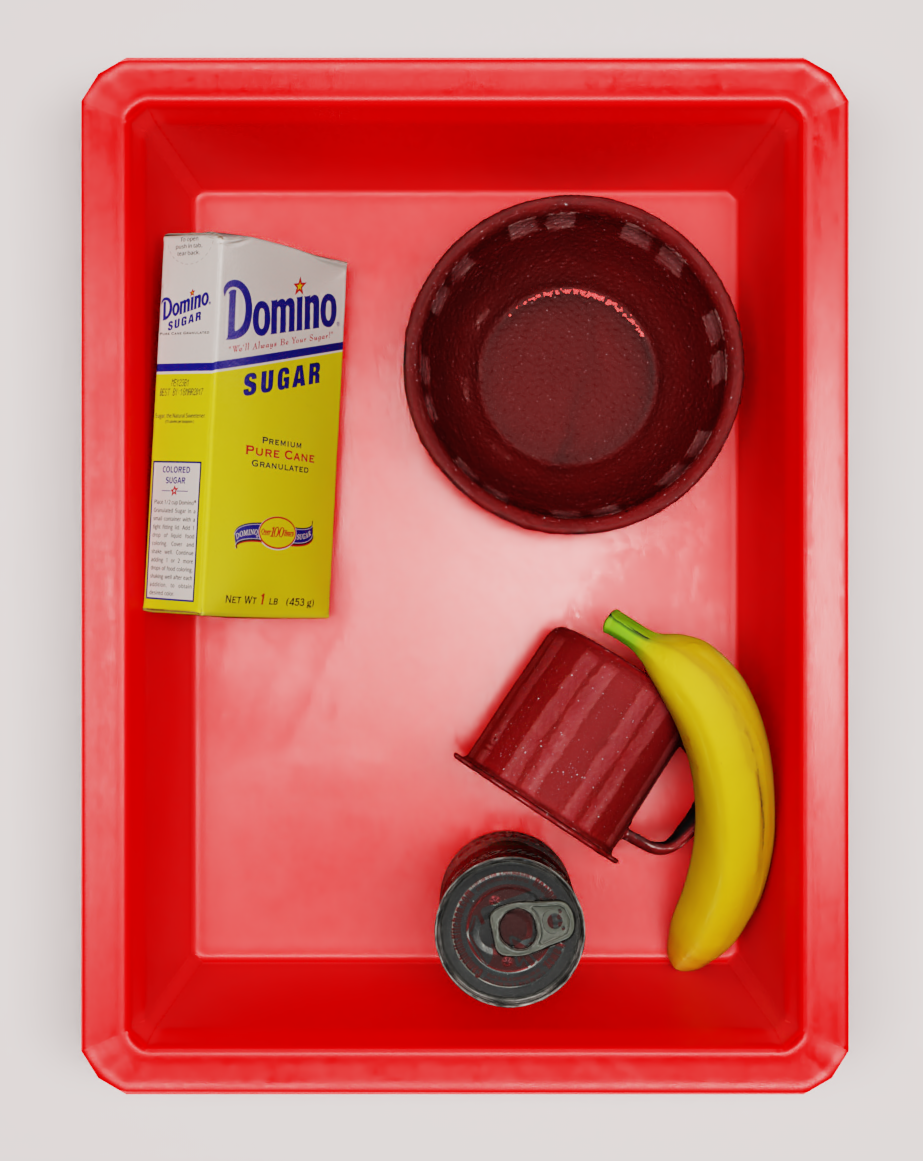,width = 0.19\textwidth}}}
    \,
	\subfloat[]{\label{fig:lvl3}{\epsfig{figure=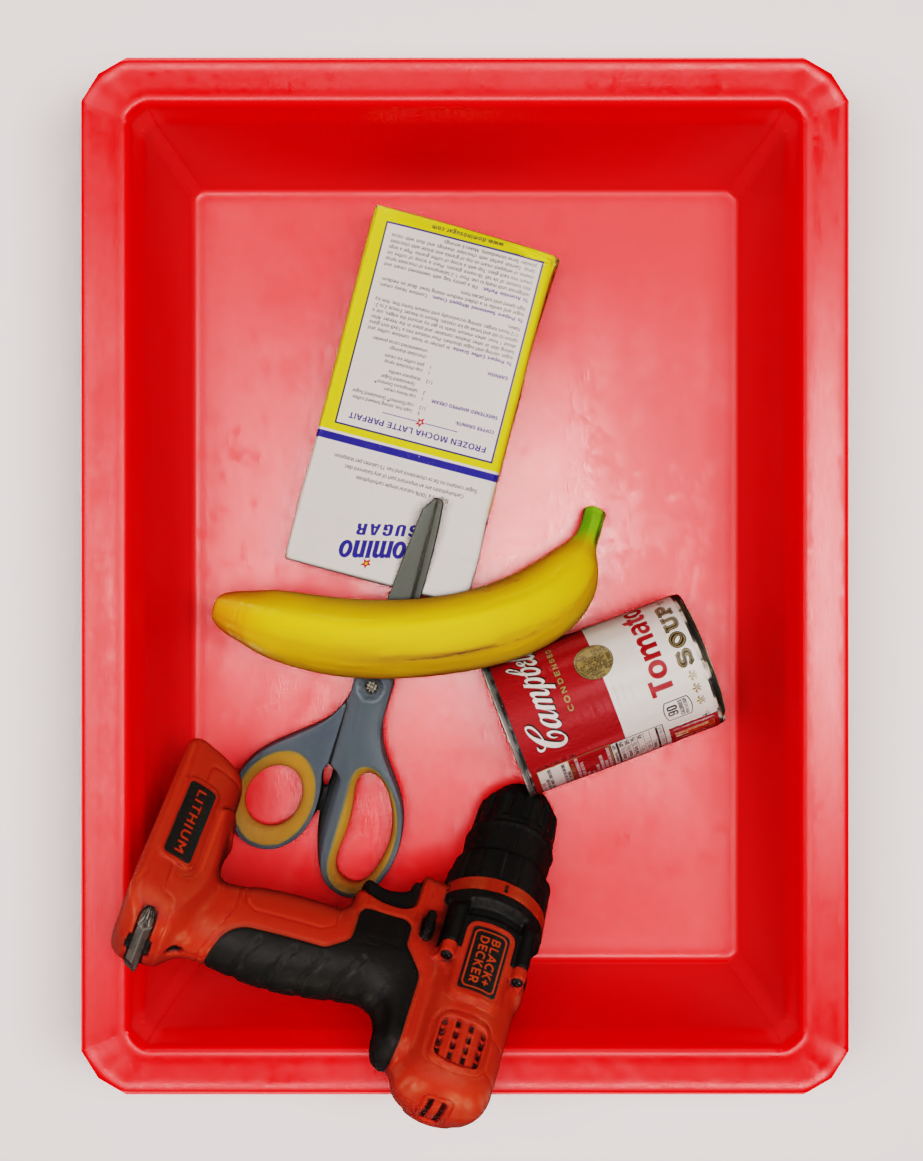,width = 0.19\textwidth}}}
	\,
	\subfloat[]{\label{fig:lvl4}{\epsfig{figure=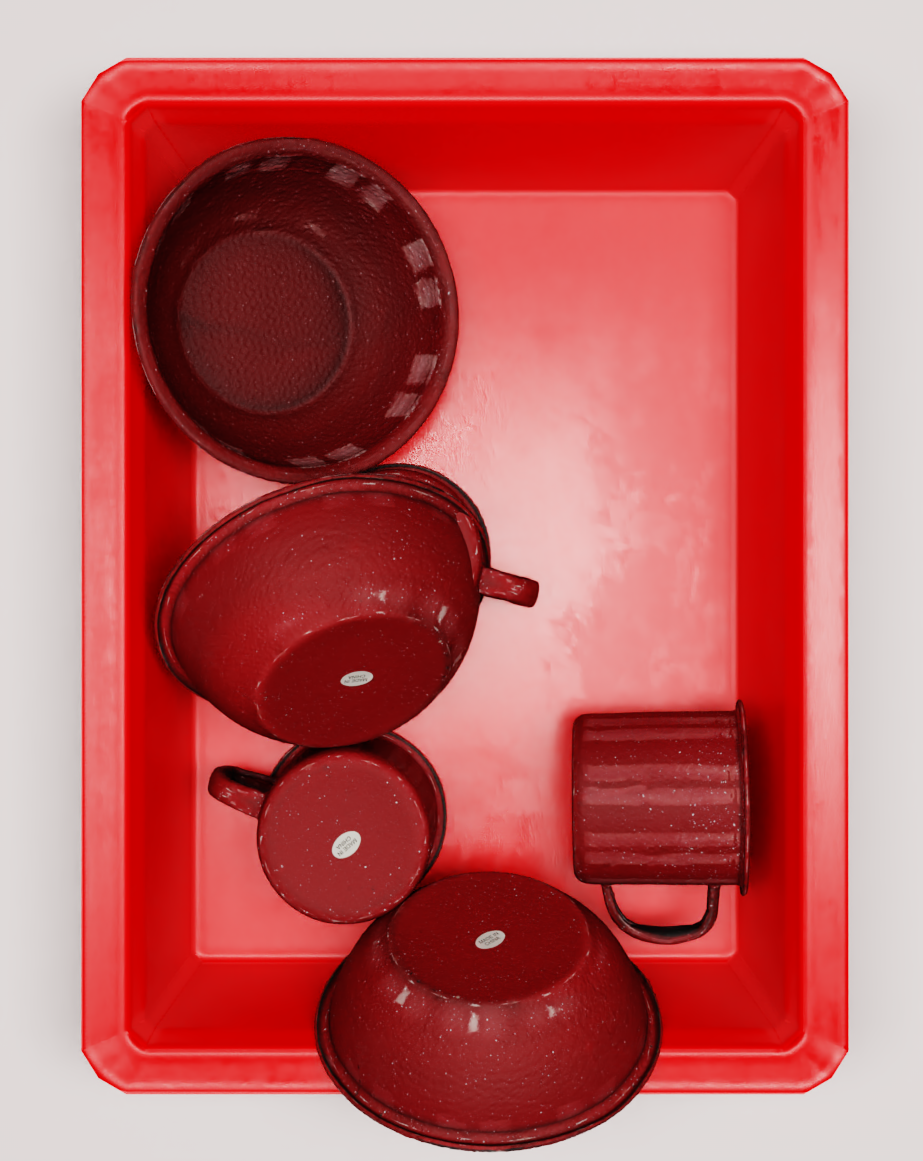,width = 0.19\textwidth}}}
	\,
	\subfloat[]{\label{fig:lvl5}{\epsfig{figure=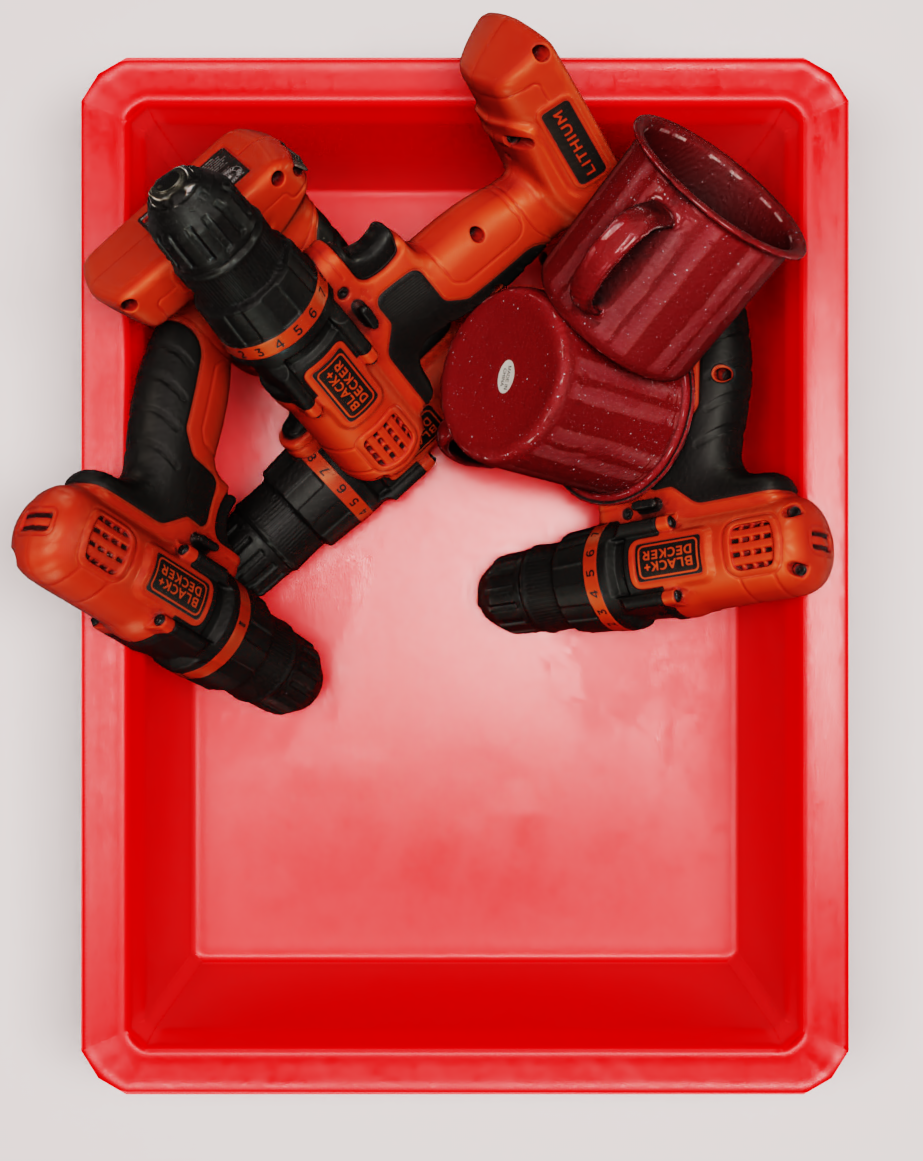,width = 0.19\textwidth}}}

	\caption{Example images from 5 difficulty levels in MetaGraspNet benchmark dataset. a) level 1: minimal occlusion. b) level 2: some occlusion. c) level 3: incomplete objects (scissors is crosscut by the banana). d) level 4: multiple instances of the same object class. e) level 5 includes all difficult characteristics. }
	\label{fig:difficulties}
\end{figure*}

The first label, occlusion percentage describes the percentage area of each object being occluded. 
This score provides an indirect measure for each object on their relationship with other objects in the layout.
This score is calculated as the percent of pixels removed in the instance segmentation mask compared to the total number of pixels of the object if all other objects in the image are removed.

% Occlusion scores can be labeled for each image by looking at percentage of each object mask remaining in the final segmentation mask label compared to the total number of pixels the object would take up on its own.
The second label is a matrix storing the relation between each pair of objects, providing a comprehensive layout representation.
To construct the relation matrix, we define three types of relationship for a pair of object $A$ and $B$. 
If $A$ is occluding $B$, we define the relationship between $(A, B)$ as positive, with a numerical value of 1.
If A is occluded by B, we define the relationship between $(A, B)$ as negative, with a numerical value of -1.
If $A$ and $B$ have no direct relationship or $A=B$, we define the relationship between $(A, B)$ as neutral, with a numerical value of 0.
Based on these definitions, for a layout with $N$ objects, we create a relation matrix with $NxN$ elements, where element $(i, j)$ in the matrix is the relationship between object $i$ and object $j$.

The third label is aiming to provide a simpler layout description in line with the robotic grasping task.
For each object in the environment, we want to answer to the following question.
How many other objects are on top of the current object that need to be moved away before picking?
To better understand the order in which objects must be grasped, we create a directed graph to represent each layout.
Each node represents an object in the layout and each edge represents an obstruction relationship where the parent object is covering the child.
From this representation, we can see what objects and how many objects are occluding the same object.
As robots pick objects sequentially, occluded objects will be revealed entirely once the objects on top of them are picked.
Therefore, it is not necessary to evaluate occluded objects that are at the bottom of the scene.
Sometimes, the occlusion between objects is small enough that can be ignored, so objects occluded by only a single other objects are also important to be evaluated. 
Given this, we categorize each object in a layout into 3 different layers.
Top layer contains objects that are clear of any obstructions.
Secondary layer includes objects that are covered by only a single other object.
Others layer includes the rest of the objects.
In some cases, there could be groups of interlocked objects.
Interlocked objects that are being directly covered by only one object would be considered to be within the secondary layer.
An example of a environment of objects from the top down view and the resulting graph can be seen in Figure \ref{fig:layers-graph-and-figure}.

\subsection{Layout-based Difficulty Levels}

While per category object detection metrics can measure performance on specific categories of objects, a difficulty rating for the overall environment would better allow us to eliminate hard examples and better understand how the model would perform under different environment conditions.

We label images according to 5 different levels of difficulty.
Those levels are defined by 4 different characteristics: Number of layers, occlusion percentage, instance completeness, and class uniqueness.
Instance completeness refers to if a single object instance is visually crosscut into multiple segments due to occlusion.
In such case, we refer to this kind of objects as incomplete objects, and refer to objects without visual crosscuts as complete objects.
Incomplete objects often result in an object over-detection or over-segmentation, and thus it is a good characteristic to test object detection and segmentation models.
Class uniqueness refers to if all objects in an image belong to different categories, or are visually distinct from each other.
This characteristic is to evaluate object detection and segmentation models on distinguishing objects with similar visual features while clustered.

The first two levels of difficulty will be primarily concerned with understanding how a model deals with different levels of occlusion and layers.
The layer limit for level 1 difficulty is set to 1, and the occlusion limit is set to 5\% empirically.
%A scene of difficulty level 1 should include no more than 2 layers (top, secondary), and contain a maximum occlusion score of ****something****
The next Three levels are primarily concerned with measuring the model's ability to correctly label object instances. Level 3 includes incomplete objects in an image, and level 4 includes non-unique objects. Level 5 includes both incomplete as well as non-unique objects.
Table \ref{table:level_description} describes all the difficulty levels, and Figure \ref{fig:difficulties} shows images from each difficulty level.

\begin{table}[htb]
\centering
    \begin{tabularx}{0.6\textwidth}{|X|X|X|X|X|}
        \hline
        Level & layer limit & occlusion limit & complete object & unique class\\ 
        \hline
         1 & 2 & 5\% & \checkmark & \checkmark\\ 
        \hline 
         2 & N/A & N/A & \checkmark & \checkmark\\
        \hline
         3 & N/A & N/A &  & \checkmark\\
        \hline
         4 & N/A & N/A & \checkmark & \\
        \hline
         5 & N/A & N/A &&\\
        \hline
    \end{tabularx}
    \caption{Table description for difficulty levels}
    \label{table:level_description}
\end{table}

\subsection{Dataset Details}
The proposed MetaGraspNet benchmark dataset contains 100,000 images, with 11,000 different scenes and 25 different household objects whose 3D models are provided by the Yale-CMU-Berkeley Object and Model Set \cite{ycb}.
The objects are placed in a red plastic box in the metaverse environment, and represents an universal small load carrier which is used in many intralogistics use-cases.
A scene is a single arrangement of objects in the bin and multiple images are taken of that scene at various viewpoints.

\section{Layout-weighted Evaluation Metric}
As discussed in Section \ref{sec:layout-label}, not all objects are of the same importance in a grasping task.
Top and secondary layer objects have a priority to be picked, while picking the rest of objects requires moving away top and secondary layer objects.
Therefore, our proposed metric focuses on evaluating top and secondary layer objects.
Besides evaluating model performance for top and secondary objects separately, we propose a layout-weighted metric which considers the model performance on both top and secondary layer objects.
In particular, we use objects' percentage of unoccluded area to weigh objects' evaluation score in each grasping scene.
Occlusion percentage measures the maximum percentage area an object could be in contact with other objects, which indirectly measures how much disturbance moving the object can cause to other objects.
The less disturbance an object creates, the more likely it is picked first, and thus the more important it is in a scene.

Given the occlusion percentage of an object, $p$, let the object's weight be $w=1-p$. 
We consider a scene $S = \{o_i| i \in [1, n]\}$ containing $n$ objects, where each object $o_i$ is indexed by $i$.
Let $T$ contain all indices for top layer objects, and $S$ contain all indices for secondary layer objects.
Let the evaluation score for object $o_i$ be $v_i$.
This score can be produced by any standard object detection and segmentation metric such as average precision, and intersection over union (IoU).
Then the per-scene layout-weighted score $V_S$ for the scene $S$ is:
\begin{equation}
    V_S = \sum_{i \in \{T+S\}}{\frac{w_i}{\sum_{j \in \{T+S\}}{w_j}}v_i}
\end{equation}
Objects from others layer are not considered during the evaluation.
Once we compute all the per-scene layout-weighted score, we take the mean as our layout-weighted score.
%Objects in the top layer are objects that can be picked up without moving other objects and thus are the most important in an object detection for robotic grasping scenario.
%One useful metric would evaluate the model against only those top objects.
%However, In a given environment, there may only be one or two objects in the top layer.
%This significantly decreases the data set we would have available for evaluation.
%As such, A second metric could also evaluate against both top and secondary objects.
%Those secondary objects could be further sorted by occlusion scores, as less occluded objects would likely result in less disturbance to other objects if picked up first.
% However, secondary objects with high occlusion would still be given priority over other objects with low occlusion.
% This is because we want to better reflect the order in which objects may be picked up, rather than simply eliminating difficult objects.

%For both situations, we would evaluate for the IOU scores of the top or top and secondary items only, as well as precision and recall scores.
%In the case of precision, predictions matching objects not in the target layers (ie. secondary, others) would not count as false positives.

%heavy occlusion leads to incorrect detection of objects

%------------------------------------------------------------------------
\section{Conclusion}
In this paper, we proposed MetaGraspNet: a large-scale benchmark dataset for vision-driven robotic grasping via physics-based metaverse synthesis.
This dataset contains 100,000 RGBD images, 11,000 scenes, and 25 classes of objects.
The proposed MetaGraspNet benchmark dataset consists of detailed object detection, segmentation, layout annotations, and a script for layout-weighted performance metric.
We presented 5 difficulties to evaluate model performance in different grasping scenarios.
Moreover, we proposed a new layout-weighted performance metric to evaluate object detection and segmentation performance in a manner that is more appropriate for robotic grasp applications.

\begin{ack}
We would like to thank the German Federal Ministry of Economic Affairs, the National Research Council Canada, Festo, and DarwinAI for supporting this work.
\end{ack}

\bibliographystyle{plainnat}  
\bibliography{references.bib}

\end{document}